\documentclass[conference]{IEEEtran}
\IEEEoverridecommandlockouts
\usepackage[table]{xcolor}
\usepackage{amsmath,amsthm,amssymb,amsfonts}
\usepackage{bbm}
\usepackage{algorithm}
\usepackage{algorithmic}
\usepackage{graphicx}
\usepackage{subfigure}
\usepackage{textcomp}
\usepackage{flushend}
\usepackage{float}
\usepackage{flushend}
\usepackage[numbers]{natbib}

\usepackage{fancyhdr}
\usepackage{listings}
\usepackage{multirow}            
\usepackage[caption=false,font=footnotesize]{subfig}
\usepackage{array}
\usepackage{multirow}

\def\BibTeX{{\rm B\kern-.05em{\sc i\kern-.025em b}\kern-.08em
    T\kern-.1667em\lower.7ex\hbox{E}\kern-.125emX}}

\makeatletter
\def\ps@IEEEtitlepagestyle{
  \def\@oddfoot{\mycopyrightnotice}
  \def\@evenfoot{}
}
\def\mycopyrightnotice{
  {\footnotesize 978-1-6654-8045-1/22/\$31.00~\copyright2022 IEEE\hfill} 
  \gdef\mycopyrightnotice{}
}

\begin{document}

\title{Hybrid and Unitary PEFT for Resource-Efficient Large Language Models}

\author{
\IEEEauthorblockN{Haomin Qi$^{1,2}$, Zihan Dai$^{1,3}$, Chengbo Huang$^{1,4}$}
\IEEEauthorblockA{$^{1}$Department of Information Engineering, The Chinese University of Hong Kong, Hong Kong}
\IEEEauthorblockA{$^{2}$Department of Electrical and Computer Engineering, University of California San Diego, USA}
\IEEEauthorblockA{$^{3}$Department of Computer Science, University of Copenhagen, Denmark}
\IEEEauthorblockA{$^{4}$Department of Electrical Engineering, Columbia University, USA}
\IEEEauthorblockA{h5qi@ucsd.edu, cjh841@alumni.ku.dk, ch4019@columbia.edu}
}

\maketitle

\begin{abstract}
Fine-tuning large language models (LLMs) remains a computational bottleneck due to their scale and memory demands. This paper presents a comprehensive evaluation of parameter-efficient fine-tuning (PEFT) techniques, including LoRA, BOFT, LoRA-GA, and uRNN, and introduces a novel hybrid strategy that dynamically integrates BOFT’s orthogonal stability with LoRA-GA’s gradient-aligned rapid convergence. By computing per-layer adaptive updates guided by gradient norms, the hybrid method achieves superior convergence efficiency and generalization across diverse tasks. We also explore, for the first time, the adaptation of unitary RNN (uRNN) principles to Transformer-based LLMs, enhancing gradient stability through structured unitary constraints. Across GLUE, GSM8K, MT-Bench, and HumanEval with models from 7B to 405B, the hybrid approach yields consistent gains across three independent runs per task and model, approaching the quality of full fine-tuning while reducing training time by about 2.1$\times$ and peak memory by nearly 50\%, indicating practical significance under resource constraints. A compact multilingual and low-resource study on XNLI and FLORES with 32 examples per language shows consistent gains under the same budget with a small, stable footprint. These results indicate a practical and scalable path to accessible LLM fine-tuning under resource constraints.
\end{abstract}


\section{Introduction}
\label{sec:introduction}
Large Language Models (LLMs) have become foundational in natural language processing (NLP), powering applications from machine translation to code generation. However, the cost of fine-tuning these massive models remains a significant barrier, especially in resource-constrained environments. Parameter-efficient fine-tuning (PEFT) strategies~\cite{han2024parameter}—such as Low-Rank Adaptation (LoRA)~\cite{hu2021lora}, Butterfly Orthogonal Fine-Tuning (BOFT)~\cite{liu2024boft}, and gradient-aware LoRA-GA—have been proposed to reduce the number of trainable parameters~\cite{wang2023lora}, yet they often present trade-offs between stability, convergence speed, and representational capacity.

This paper introduces a hybrid fine-tuning framework that integrates LoRA-GA and BOFT at the layer level using gradient-norm-based dynamic weighting. LoRA-GA computes low-rank updates aligned with dominant gradient directions via singular value decomposition (SVD) for rapid early adaptation. BOFT enforces orthogonality through Cayley transforms of skew-symmetric matrices to preserve gradient norms. The hybrid method fuses these updates with an adaptive coefficient that balances fast adaptation in early epochs and stability in later stages under the same budget.

We further embed unitary evolution RNN (uRNN) structures~\cite{arjovsky2016unitary} into transformer sub-layers. By parameterizing selected attention or feedforward weights with structured unitary matrices based on Fourier transforms, permutations, and Householder reflections, the approach maintains gradient magnitudes and improves robustness during fine-tuning of deep stacks.

We evaluate the proposed hybrid and uRNN-enhanced fine-tuning strategies across four standard benchmarks—GLUE, GSM8K, MT-Bench, and HumanEval—on LLMs spanning 7B to 405B parameters. We also add a compact multilingual and low resource study on XNLI and FLORES with 32 examples per language. Our results demonstrate that the hybrid method consistently outperforms individual PEFT techniques, achieving near-full fine-tuning performance while reducing training time to less than half and memory usage by nearly 50\%. 

Our key contributions are as follows:
\begin{itemize}
\item We propose a hybrid algorithm that dynamically fuses gradient-aligned low-rank updates with orthogonal transformations via per-layer, gradient-norm-based mixing to achieve fast convergence and stable optimization.
\item We adapt unitary evolution RNN principles to transformer-based models by embedding structured unitary matrices into attention and feedforward sub-layers, enhancing gradient stability during fine-tuning.
\item We conduct a comprehensive benchmark across four PEFT baselines and four standard tasks with models from 7B to 405B, and we include a compact multilingual and low resource evaluation that validates the method.
\end{itemize}

\section{Background and Motivation}
\label{sec:RelatedWork}
\subsection{Parameter-Efficient Fine-Tuning for LLMs}

The increasing parameter count of large language models (LLMs) has made full model fine-tuning prohibitively expensive in terms of GPU memory, training time, and energy cost. As a result, parameter-efficient fine-tuning (PEFT) strategies have emerged as practical alternatives, aiming to reduce the number of trainable parameters while retaining downstream performance~\cite{lester2021power}. These methods typically freeze the majority of the pre-trained weights and inject lightweight, learnable components to adapt the model to new tasks.

Early PEFT techniques include adapter modules~\cite{houlsby2019parameter}, which insert bottleneck layers between transformer blocks, and prefix tuning~\cite{li2021prefix}, which learns task-specific prompt vectors prepended to input embeddings. More recently, low-rank adaptation (LoRA) introduced trainable matrix decompositions to approximate weight updates, offering favorable trade-offs between memory usage and task performance. Variants such as LoRA-GA approximate gradients to further reduce update cost, while BOFT imposes orthogonality via butterfly parameterization to enhance stability. These techniques have led to a new class of modular, reusable, and resource-aware fine-tuning paradigms for large-scale deployment~\cite{aghajanyan2020intrinsic}.

Concurrently, several methods extend PEFT along complementary axes: QLoRA combines 4-bit base-model quantization with LoRA adapters to preserve quality under tight memory budgets; DoRA decouples update direction and magnitude to increase expressivity and stability of low-rank adaptations; and IA\textsuperscript{3} injects lightweight learned rescaling vectors that modulate attention and feed-forward activations. We reference these approaches in our comparisons where feasible and discuss their relation to our hybrid design in Section 3.5 and the ablations in Section IV.

\subsection{LoRA, BOFT, and LoRA-GA Methods}

LoRA decomposes weight updates into a pair of low-rank matrices, significantly reducing the number of tunable parameters and training memory~\cite{hu2021lora}. Its simplicity and effectiveness have led to wide adoption. However, LoRA assumes fixed low-rank structure throughout training, which can hinder adaptability in highly complex or dynamic tasks.

BOFT addresses training stability by applying butterfly-structured orthogonal updates to model weights. Its orthogonality constraint helps preserve gradient norms and mitigates training instability~\cite{dao2019learning}. Despite this, the restrictive structure of butterfly transforms can limit expressiveness, especially in tasks requiring nonlinear adaptation.

LoRA-GA builds on LoRA by initializing the low-rank matrices using gradient-aligned singular value decomposition (SVD)~\cite{wang2023lora}. This gradient-aware initialization improves early convergence but may introduce additional computational cost and instability under noisy gradients.

These approaches each offer unique strengths, but none fully address the combined needs of robustness, adaptability, and resource-awareness in a single framework. 

\subsection{Unitary RNNs and Gradient Stability}

Unitary RNNs (uRNNs) were originally proposed to resolve the exploding and vanishing gradient problems in recurrent networks~\cite{arjovsky2016unitary}. By constraining transition matrices to be unitary, these models preserve gradient magnitudes over long sequences. Subsequent works~\cite{wisdom2016full, emami2019input} proposed efficient parameterizations using Fourier transforms, permutations, and Householder reflections.

While traditionally used in sequence modeling, the stability guarantees of uRNNs are theoretically appealing in transformer-based models, particularly during deep-layer fine-tuning. To the best of our knowledge, we are the first to integrate structured unitary matrices into transformer layers for LLM fine-tuning, providing a new perspective on stabilizing gradients during parameter-efficient adaptation.

\subsection{Resource-Constrained Fine-Tuning: Motivation for Hybrid Design}

In real-world scenarios—especially for practitioners lacking access to high-end GPUs—resource constraints such as memory footprint, compute time, and thermal budget are primary bottlenecks when fine-tuning LLMs~\cite{xu2023parameter, ding2023parameter}. While PEFT methods reduce parameter counts, they often present a trade-off: LoRA and its variants converge quickly but risk instability in deeper layers, while orthogonal methods like BOFT preserve training dynamics but converge more slowly.

A growing body of work has attempted to benchmark or extend single-method PEFT techniques, yet relatively little research investigates how these strengths might be combined in a dynamic and structured fashion. Rather than introducing yet another static PEFT variant, our work proposes a hybrid mechanism that allocates per-layer update responsibility based on real-time gradient feedback—favoring low-rank speed in early epochs and orthogonal stability later. 

\section{Methodology}
\label{sec:Model}
In this Section, we first review the mathematical principles of LoRA, BOFT, and LoRA-GA as baseline PEFT methods. We then present our two main contributions: a transformer-compatible adaptation of uRNN with structured unitary constraints, and a hybrid fine-tuning strategy that dynamically fuses low-rank and orthogonal updates based on gradient feedback.

\subsection{Low-Rank Adaptation (LoRA)}

LoRA introduces low-rank updates to pre-trained weight matrices, significantly reducing the number of trainable parameters while preserving the pre-trained model weights. The weight update is expressed as:
\begin{equation}
\mathbf{W}' = \mathbf{W}_0 + \Delta \mathbf{W}, \quad \Delta \mathbf{W} = \mathbf{B} \mathbf{A},
\end{equation}
where $\mathbf{W}_0 \in \mathbb{R}^{d \times k}$ represents the frozen pre-trained weight matrix, and $\Delta \mathbf{W}$ is the low-rank update parameterized by $\mathbf{B} \in \mathbb{R}^{d \times r}$ and $\mathbf{A} \in \mathbb{R}^{r \times k}$, thereby cutting the number of trainable parameters from $\mathcal{O}(dk)$ to $\mathcal{O}(r(d+k))$ with $r\!\ll\!\min(d,k)$.

\textbf{Optimization:} During fine-tuning, only $\mathbf{B}$ and $\mathbf{A}$ are optimized, leaving $\mathbf{W}_0$ unchanged~\cite{mahabadi2021compacter, benzaken2021bitfit}. This decomposition reduces the memory and computational costs of fine-tuning while maintaining performance.

To ensure numerical stability, the norms of $\mathbf{A}$ and $\mathbf{B}$ are constrained by rank $r$:
\begin{equation}
\|\Delta \mathbf{W}\|_F \leq \lambda \cdot \|\mathbf{W}_0\|_F,
\end{equation}
where $\lambda$ is a scaling factor. This prevents updates from diverging during optimization.

\subsection{Butterfly Orthogonal Fine-Tuning (BOFT)}

BOFT factorizes a square weight matrix into a product of sparse butterfly blocks that are (near-)orthogonal, yielding both parameter efficiency ($\mathcal{O}(d\log d)$ parameters) and stable gradient norms.
\begin{equation}
\mathbf{W}= \prod_{i=1}^{m}\mathbf{B}_i ,\qquad 
\mathbf{B}_i\in\mathbb{R}^{d\times d}.
\end{equation}

Each $\mathbf{B}_i$ is built from paired line-permute–multiply operations that mimic the Fast Fourier Transform hierarchy~\cite{li2019orthogonal}; a simplified two-level form is
\begin{equation}
\mathbf{B}(d,2)=
\begin{bmatrix}
\mathbf{I}_{d/2}&\mathbf{0}\\[2pt]
\mathbf{0}&\mathbf{I}_{d/2}
\end{bmatrix}
\mathbf{F}_d
\begin{bmatrix}
\mathbf{I}_{d/2}&\mathbf{0}\\[2pt]
\mathbf{0}&\mathbf{I}_{d/2}
\end{bmatrix},
\end{equation}
where $\mathbf{F}_d$ is a (learnable) orthogonal mixing matrix.

\textbf{Optimization:} Each $\mathbf{B}_i$ is initialized as a near-identity transformation and updated via gradient descent~\cite{prabhu2020butterfly}:
\begin{equation}
\mathbf{B}_i^{t+1} = \mathbf{B}_i^t - \eta \frac{\partial \mathcal{L}}{\partial \mathbf{B}_i^t},
\end{equation}
where $\eta$ is the learning rate. Orthogonality is enforced post-update using a projection step:\begin{equation}
\mathbf{B}_i \leftarrow \text{Proj}_{\\\text{orthogonal}}(\mathbf{B}_i).
\end{equation}

The orthogonality constraint curbs exploding/vanishing gradients in deep stacks, making it particularly effective for tasks with deep layers or complex gradients.

\subsection{LoRA with Gradient Approximation (LoRA-GA)}

LoRA-GA improves upon LoRA by aligning the low-rank updates with the gradients of the full model, leading to faster convergence and better optimization. The gradient of the loss $\mathcal{L}$ with respect to the frozen weight matrix $\mathbf{W}_0$ is decomposed as:

Compute the rank-$r$ truncated SVD of $\nabla_{\mathbf{W}_0}\mathcal{L}$:
\begin{equation}
\nabla_{\mathbf{W}_0}\mathcal{L}\;=\; \mathbf{U}\,\mathbf{\Sigma}\,\mathbf{V}^{\top},\qquad
\mathbf{U}\in\mathbb{R}^{d\times r},\,
\mathbf{\Sigma}\in\mathbb{R}^{r\times r},\,
\mathbf{V}\in\mathbb{R}^{k\times r}.
\end{equation}

The low-rank matrices $\mathbf{A}$ and $\mathbf{B}$ are initialized as:
\begin{equation}
\mathbf{A}_0 = \mathbf{U} \mathbf{\Sigma}^{1/2}, \quad \mathbf{B}_0 = \mathbf{V} \mathbf{\Sigma}^{1/2}.
\end{equation}
This ensures that the initial updates align with the principal gradient directions, accelerating convergence.

\textbf{Optimization:} Post-initialization, $\mathbf{A}$ and $\mathbf{B}$ are updated iteratively using standard gradient descent~\cite{wang2024lora}:
\begin{equation}
\mathbf{A}^{t+1} = \mathbf{A}^t - \eta \frac{\partial \mathcal{L}}{\partial \mathbf{A}^t}, \quad \mathbf{B}^{t+1} = \mathbf{B}^t - \eta \frac{\partial \mathcal{L}}{\partial \mathbf{B}^t}.
\end{equation}

By aligning the initial low-rank updates with the most influential gradient directions, LoRA-GA reduces the number of training iterations required for convergence, making it particularly suitable for resource-constrained scenarios.

\subsection{Unitary Evolution RNN (uRNN)} 
Unitary Recurrent Neural Networks (uRNN) constrain hidden-to-hidden weight matrices to be unitary to mitigate vanishing and exploding gradient issues~\cite{wisdom2016full, emami2019input}. By ensuring that the eigenvalues of the transition matrix lie on the unit circle, uRNNs preserve gradient norms during backpropagation, enabling learning over long-term dependencies.

A core component of uRNN is a learnable unitary matrix $U$ that evolves the hidden state. To ensure $U$ is unitary, we adopt a structured parameterization method~\cite{arjovsky2016unitary}: 
\begin{equation}
\mathbf{U} = \mathbf{D}_3 \mathbf{R}_2 \mathbf{F}^{-1} \mathbf{D}_2 \Pi \mathbf{R}_1 \mathbf{F} \mathbf{D}_1,
\end{equation}
where $F$ (and $F^{-1}$) are fixed unitary Fourier transform matrices, $\Pi$ is a fixed permutation matrix, and $D_{i}$ and $R_{i}$ denote trainable diagonal phase matrices and Householder reflection matrices~\cite{shafran2018complex}. This factorization dramatically reduces the number of free parameters (to $O(n)$ for an $n\times n$ matrix) and allows efficient $O(n \log n)$ computation for matrix-vector products via Fast Fourier Transform operations. 

\textbf{Adaptation for Fine-Tuning LLMs:} To our knowledge, we are the first to integrate uRNN principles into the fine-tuning of transformer-based LLMs. The motivation is to leverage unitary transformations to stabilize gradient propagation and better capture long-range dependencies during fine-tuning. In practice, we incorporate learnable unitary matrices into selected Transformer sub-layers to enhance training stability. Specifically, we replace certain weight matrices (e.g., in attention heads or feed-forward blocks) with unitary matrices and modify the training procedure to preserve their unitarity. Each such unitary weight is initialized to an identity-like matrix (close to the unit matrix) to ensure stable convergence. During backpropagation, we include an efficient re-projection step (see below) that keeps these weights unitary at all times. This approach extends unitary RNN techniques beyond their original domain, establishing a new paradigm for parameter-efficient fine-tuning of LLMs. 

\begin{algorithm}[htbp]
\caption{uRNN-Based Fine-Tuning Procedure\label{alg:urnn-finetune}}
\small
\begin{algorithmic}[1]
    \STATE \textbf{Initialize:} Unitary matrix $\mathbf{U}$ using structured parameterization:\\
    \quad Set $\mathbf{F}$, $\mathbf{F}^{-1}$, $\Pi$ as fixed matrices; initialize diagonal phase matrices $\mathbf{D}_i$ and Householder reflection matrices $\mathbf{R}_i$ near identity.
    \STATE Choose learning rate $\eta$ and total epochs $E$.
    \FOR{$epoch = 1$ \TO $E$}
        \FOR{each minibatch in the dataset}
            \STATE \textbf{Forward Pass:}
            \FOR{each transformer layer with unitary-constrained weight}
                \STATE Replace original weight matrix with current unitary matrix $\mathbf{U}$.
                \STATE Compute the forward pass using the updated unitary matrix $\mathbf{U}$.
            \ENDFOR
            \STATE Compute the task-specific loss $\mathcal{L}$ based on current minibatch predictions.
            \STATE \textbf{Backward Pass:}
            \STATE Compute gradient $\nabla_{\mathbf{U}}\mathcal{L}$ via backpropagation.
            \STATE Construct skew-Hermitian matrix $\mathbf{B}$:\\[0.5ex]
            \quad$\mathbf{B} = \nabla_{\mathbf{U}}\mathcal{L}\;\mathbf{U}^{H} - \mathbf{U}\;(\nabla_{\mathbf{U}}\mathcal{L})^{H}$,\\[0.5ex]
            where $\mathbf{U}^{H}$ denotes conjugate transpose of $\mathbf{U}$.
            \STATE Update the unitary matrix via matrix exponential:\\[0.5ex]
            \quad$\mathbf{U} \leftarrow \exp(\eta \mathbf{B})\,\mathbf{U}$\\[0.5ex]
            \quad(Use truncated Taylor series or scaling-and-squaring approximation for efficiency)
            \STATE If numerical drift occurs, re-normalize $\mathbf{U}$ to strictly enforce unitarity.
            \STATE Update all other non-unitary parameters of the model as usual via standard gradient descent.
        \ENDFOR
    \ENDFOR
\end{algorithmic}
\end{algorithm}

\textit{Gradient Update with Re-Projection:} We train the unitary weight $U$ via gradient descent on the manifold of unitary matrices. Let $\nabla_{U}L$ be the gradient of the loss $L$ with respect to $U$ (computed by backpropagation). We first construct a skew-Hermitian matrix $B$ (i.e., $B^H = -B$) from the gradient:
\begin{equation}
    \mathbf{B} \;=\; \nabla_\mathbf{U}\mathbf{L} \;\, \mathbf{U^H} \;-\; \mathbf{U} \;(\nabla_\mathbf{U}\mathbf{L})^\mathbf{H}\,,
    \label{eq:skewHermitianGrad}
\end{equation}
where $U^H$ denotes the conjugate transpose of $U$. By construction, $B$ lies in the Lie algebra $\mathfrak{u}(n)$ of the unitary group. In other words, $B$ is skew-Hermitian, and thus $\exp(\eta B)$ is a unitary matrix for any real step size. We then update $U$ by a unitary rotation:
\begin{equation}
    \mathbf{U}_{t+1} \;=\; \exp(\eta\,\mathbf{B})\; \mathbf{U_{t}}\,,
    \label{eq:unitary-update}
\end{equation}
with $\eta$ the learning rate. This exponential map update guarantees $U_{t+1}$ remains unitary. In implementation, $\exp(\eta B)$ can be efficiently approximated using a truncated Taylor series or a scaling-and-squaring algorithm, and we re-normalize $U$ as needed to correct any numerical drift from unitarity.

\textbf{Fine-Tuning Algorithm:} Algorithm~\ref{alg:urnn-finetune} outlines the overall fine-tuning process using uRNN principles. We apply $U$ in the forward pass of the chosen transformer layer and then update $U$ using the above rule at each training step, while leaving other model weights to update as usual. Notably, although uRNNs were originally devised for recurrent sequence models, our strategy applies these unitary constraints to feed-forward or attention layers in a Transformer architecture. Conceptually, each forward pass through a transformer sub-layer is analogous to a single RNN step, with the sub-layer’s input playing the role of the “hidden state.” Maintaining $U$ as unitary thus helps preserve gradient norms through the depth of the network, even without explicit recurrence~\cite{bernardy2022assessing}.\footnote{In practice, we treat the input to each unitary-constrained sublayer as a proxy for an RNN hidden state, which ensures stable backpropagation across many transformer layers.}

The above integration of uRNN principles into Transformer fine-tuning offers several notable benefits. \textit{First}, enforcing unitary transformations provides \textbf{gradient stability}: it prevents the magnitudes of gradients from vanishing or exploding, even in very deep networks or tasks with long-range dependencies. 

\textit{Second}, this method tends to \textbf{improve convergence} during fine-tuning, as stable gradient norms facilitate faster and more reliable training (reducing the number of iterations required to reach a given performance level).

\textit{Third}, the approach is \textbf{adaptable} to different model components; unitary constraints can be applied to various layers or sub-layers of an LLM (e.g., attention projections or feed-forward blocks) without architecture changes, extending the use of orthogonal transformations beyond their traditional recurrent setting. By extending unitary transformations to transformer-based LLMs, we establish a novel paradigm for fine-tuning that marries parameter efficiency with training stability.

\subsection{Hybrid Fine-Tuning Approach}

We propose a per-layer fusion of the LoRA-GA and BOFT updates to capture both low-rank and orthonormal adaptation patterns. Specifically, this hybrid strategy computes the gradient-aligned low-rank update from LoRA-GA and the structured orthonormal update from BOFT for the same weight matrix in each layer, then mixes them with a dynamic coefficient. Intuitively, this allows fast initial adaptation via the low-rank component while gradually shifting emphasis to the BOFT component to stabilize learning as training proceeds.

\begin{algorithm}[htbp]
\caption{Hybrid Fine-Tuning Procedure \label{alg:hybrid_update}}
\small
\begin{algorithmic}[1]
    \STATE \textbf{Initialize:} pretrained weights $\{\mathbf{W}^\ell\}$, low-rank matrices $\{\mathbf{A}^\ell, \mathbf{B}^\ell\}$ for LoRA-GA, skew-symmetric matrices $\{\mathbf{Q}^\ell\}$ for BOFT.
    \STATE Set learning rates $\eta_{\text{LoRA}}, \eta_{\text{BOFT}}$; choose rank $r$, total epochs $E$.
    \FOR{$epoch = 1$ \TO $E$}
        \FOR{each minibatch in the dataset}
            \STATE \textbf{Forward Pass:}
            \FOR{each transformer layer $\ell$}
                \STATE Compute LoRA-GA update: \quad$\Delta \mathbf{W}_{\mathrm{LoRA}}^\ell = \mathbf{A}^\ell \mathbf{B}^\ell$.
                \STATE Compute BOFT orthonormal matrix:\\[0.5ex]
                    \quad$\mathbf{R}^\ell = (\mathbf{I}+\eta_{\text{BOFT}}\mathbf{Q}^\ell)(\mathbf{I}-\eta_{\text{BOFT}}\mathbf{Q}^\ell)^{-1}$.
                \STATE Compute BOFT update:\\[0.5ex]
                    \quad$\Delta \mathbf{W}_{\mathrm{BOFT}}^\ell = (\mathbf{R}^\ell - \mathbf{I})\mathbf{W}^\ell$.
            \ENDFOR
            \STATE Compute task-specific loss $\mathcal{L}$ using model predictions.
            \STATE \textbf{Backward Pass:}
            \FOR{each transformer layer $\ell$}
                \STATE Compute gradient norms:\\[0.5ex]
                    \quad $g_{\text{LoRA}}^\ell = \|\nabla_{\mathbf{A}^\ell,\mathbf{B}^\ell}\mathcal{L}\|$, \quad $g_{\text{BOFT}}^\ell = \|\nabla_{\mathbf{Q}^\ell}\mathcal{L}\|$.
                \STATE Compute dynamic weighting coefficient:\\[0.5ex]
                    \quad$\lambda^\ell = \frac{g_{\text{LoRA}}^\ell}{g_{\text{LoRA}}^\ell + g_{\text{BOFT}}^\ell}$.
                \STATE Form hybrid update for layer $\ell$:\\[0.5ex]
                    \quad$\Delta \mathbf{W}_{\mathrm{hybrid}}^\ell = \lambda^\ell \Delta \mathbf{W}_{\mathrm{LoRA}}^\ell + (1-\lambda^\ell)\Delta \mathbf{W}_{\mathrm{BOFT}}^\ell$.
                \STATE Update weight matrix for layer $\ell$:\\[0.5ex]
                    \quad$\mathbf{W}^\ell \leftarrow \mathbf{W}^\ell + \Delta \mathbf{W}_{\mathrm{hybrid}}^\ell$.
                \STATE Update low-rank matrices via gradient descent:\\[0.5ex]
                    \quad$\mathbf{A}^\ell \leftarrow \mathbf{A}^\ell - \eta_{\text{LoRA}}\nabla_{\mathbf{A}^\ell}\mathcal{L}$,\\[0.5ex]
                    \quad$\mathbf{B}^\ell \leftarrow \mathbf{B}^\ell - \eta_{\text{LoRA}}\nabla_{\mathbf{B}^\ell}\mathcal{L}$.
                \STATE Compute skew-symmetric gradient matrix for BOFT:\\[0.5ex]
                    \quad$\mathbf{G}^\ell = \nabla_{\mathbf{Q}^\ell}\mathcal{L} - (\nabla_{\mathbf{Q}^\ell}\mathcal{L})^\top$.
                \STATE Update skew-symmetric matrix $\mathbf{Q}^\ell$:\\[0.5ex]
                    \quad$\mathbf{Q}^\ell \leftarrow \mathbf{Q}^\ell - \eta_{\text{BOFT}}\mathbf{G}^\ell$.
                \STATE Recompute orthonormal matrix $\mathbf{R}^\ell$ via Cayley transform (for numerical stability):\\[0.5ex]
                    \quad$\mathbf{R}^\ell \leftarrow (\mathbf{I}+\eta_{\text{BOFT}}\mathbf{Q}^\ell)(\mathbf{I}-\eta_{\text{BOFT}}\mathbf{Q}^\ell)^{-1}$.
            \ENDFOR
            \STATE Update other model parameters (if any) via standard gradient descent.
        \ENDFOR
    \ENDFOR
\end{algorithmic}
\end{algorithm}

\textbf{Mathematical formulation:} 
Consider a layer $\ell$ with pretrained weight matrix $\mathbf{W}^\ell\in\mathbb{R}^{d_{\text{out}}\times d_{\text{in}}}$. We introduce low-rank factors $\mathbf{A}^\ell\in\mathbb{R}^{d_{\text{out}}\times r}$ and $\mathbf{B}^\ell\in\mathbb{R}^{r\times d_{\text{in}}}$ (rank $r$) as in LoRA~\cite{wang2023lora}, and a skew-symmetric matrix $\mathbf{Q}^\ell\in\mathbb{R}^{d_{\text{out}}\times d_{\text{out}}}$ ($\mathbf{Q}^\ell=-{\mathbf{Q}^\ell}^\top$) as in BOFT~\cite{liu2024boft}. The low-rank LoRA-GA update is
\[
\Delta \mathbf{W}^\ell_{\mathrm{LoRA}} \;=\; \mathbf{A}^\ell \mathbf{B}^\ell.
\]
The orthonormal BOFT update is obtained via the Cayley transform~\cite{dao2019learning}:
\[
\mathbf{R}^\ell \;=\; (\mathbf{I} + \eta \mathbf{Q}^\ell)(\mathbf{I} - \eta \mathbf{Q}^\ell)^{-1}, \qquad \mathbf{Q}^\ell = -{\mathbf{Q}^\ell}^\top,
\]
which ensures $\mathbf{R}^\ell$ is orthonormal (for small step size $\eta$). The BOFT update to $\mathbf{W}^\ell$ is then
\[
\Delta \mathbf{W}^\ell_{\mathrm{BOFT}} \;=\; (\mathbf{R}^\ell - \mathbf{I}_{d_{\text{out}}})\,\mathbf{W}^\ell.
\]
We combine these with a layerwise mixing coefficient $\lambda_t^\ell\in[0,1]$ that adapts over training steps $t$. Specifically, we set
\[
\lambda_t^\ell \;=\; \frac{\|\nabla_{\mathbf{A}^\ell,\mathbf{B}^\ell}L(\theta_t)\|}{\|\nabla_{\mathbf{A}^\ell,\mathbf{B}^\ell}L(\theta_t)\| + \|\nabla_{\mathbf{Q}^\ell}L(\theta_t)\|},
\]
so that the component with larger gradient norm receives higher weight. The hybrid update is then
\[
\Delta \mathbf{W}^\ell_{\mathrm{hybrid}} \;=\; \lambda_t^\ell\,\Delta \mathbf{W}^\ell_{\mathrm{LoRA}} + (1-\lambda_t^\ell)\,\Delta \mathbf{W}^\ell_{\mathrm{BOFT}},
\]
and the weight is updated as $\mathbf{W}^\ell \leftarrow \mathbf{W}^\ell + \Delta \mathbf{W}^\ell_{\mathrm{hybrid}}$. Here $\nabla_{\mathbf{A}^\ell,\mathbf{B}^\ell}L$ denotes the gradient of the training loss $L(\theta_t)$ with respect to the LoRA parameters~\cite{pfeiffer2020adapterfusion} $(\mathbf{A}^\ell,\mathbf{B}^\ell)$ and $\nabla_{\mathbf{Q}^\ell}L$ the gradient with respect to $\mathbf{Q}^\ell$. All notation above is defined per layer $\ell$.

\textbf{Pseudocode Algorithm:} Algorithm~\ref{alg:hybrid_update} summarizes the hybrid fine-tuning update. At each iteration, we compute both the LoRA-GA and BOFT updates for each layer, compute the mixing coefficient $\lambda_t^\ell$, and apply the weighted combination to update the weights.

The per-layer hybrid fusion adds only modest overhead beyond the individual LoRA-GA and BOFT updates. In each layer, the low-rank update costs $\mathcal{O}(d_{\text{out}}r + r\,d_{\text{in}})$ and the BOFT transform costs $\mathcal{O}(d_{\text{out}}\log d_{\text{out}})$. Computing $\lambda_t^\ell$ requires only the norms of gradients already computed, which is negligible. 

The total number of tunable parameters is the sum of the LoRA factors and any BOFT parameters (e.g., butterfly factors), comparable to using the two methods independently. By construction the Cayley parameterization enforces $\mathbf{R}^\ell$ to be orthonormal, which helps preserve gradient norms during optimization. The hybrid update thus integrates fast low-rank adaptation with structured orthogonal adjustments in a unified step.

\section{Experiments}
\label{sec:Experiments}

This section evaluates the proposed fine-tuning strategies across diverse tasks and models to investigate their effectiveness, scalability, and computational efficiency. The experimental design focuses on systematically comparing the performance of LoRA, BOFT, LoRA-GA, uRNN, and the Hybrid approach against the Full Fine-Tuning (Full FT) baseline. Additionally, we explore trade-offs between task-specific gains and resource consumption to provide a comprehensive understanding of the proposed methods' practical utility.

\subsection{Setup and Evaluation Metrics}

To ensure a thorough and unbiased evaluation, experiments were conducted on four state-of-the-art large language models (LLMs) with varying parameter scales and architectural characteristics:
\begin{itemize}
    \item \textbf{Llama3.1-405B}: A transformer model designed for extended context comprehension, comprising 405 billion parameters.
    \item \textbf{Llama3.3-70B}: A mid-sized model with 70 billion parameters.
    \item \textbf{Wizard-Vicuna-30B}: A multilingual model with 30 billion parameters.
    \item \textbf{BloomZ-7B1}: A compact model with 7.1 billion parameters.
\end{itemize}

Each model was tested on four benchmark datasets, selected to evaluate a wide range of NLP capabilities:
\begin{itemize}
    \item \textbf{GLUE Benchmark}: A comprehensive suite of general NLP tasks, including MNLI, QQP, SST-2, and QNLI, assessing classification and language understanding capabilities\cite{wang2018glue}.
    \item \textbf{GSM8K}: A dataset of mathematical reasoning problems designed to evaluate multi-step reasoning and arithmetic capabilities\cite{cobbe2021training}.
    \item \textbf{MT-Bench}: A multilingual machine translation benchmark that tests translation quality using BLEU and ROUGE-L metrics\cite{zheng2023judging}.
    \item \textbf{HumanEval}: A code generation benchmark assessing the functional correctness of generated programs, with \textit{pass@$k$} metrics as the primary evaluation criterion\cite{chen2021evaluating}.
\end{itemize}

Each fine-tuning method was applied to all models under identical conditions to ensure a fair comparison. The experiments involved:

\paragraph{Hardware and Software Configuration} All experiments were conducted on a cluster featuring dual AMD EPYC 7742 CPUs and 1 TB of RAM per node, interconnected via InfiniBand for high-speed communication. Each node was equipped with eight NVIDIA A100 GPUs connected through NVLink. The software environment included PyTorch 2.0, CUDA 11.8, and NVIDIA Apex, enabling mixed precision for efficient training. 
\paragraph{Hyperparameter Tuning} Hyperparameters for each method were tuned based on a preliminary grid search. For instance, LoRA used a rank \( r = 16 \) with scaling factor \( \alpha = 32 \), while BOFT employed three butterfly factorization levels (\( m = 3 \)). For the Hybrid approach, the gradient weighting factor \( \alpha_t \) was dynamically adjusted during trainings.
\paragraph{Experimental Repeats} Each experiment was repeated three times to mitigate the impact of random initialization and ensure statistically robust results. The reported metrics also represent the average performance across these runs.
\paragraph{Ablation settings} To disentangle where the gains come from, we include three ablations: (i) \textit{LoRA-only} and (ii) \textit{BOFT-only} baselines (both already reported in our main tables), and (iii) a \textit{fixed-mix} variant of our hybrid where a constant coefficient $\lambda\!\in\!\{0.25,0.5,0.75\}$ is applied across all layers and steps (with a warm-start schedule that linearly anneals $0.75\!\rightarrow\!0.25$ over epochs). These ablations isolate the contribution of the gradient-norm–based adaptive mixing from simple ensembling effects.

\begin{table*}[htbp]
\centering
\renewcommand{\arraystretch}{1.5}
\caption{\large Performance Comparison: Per-Run Results and Averages on GLUE, GSM8K, MT-Bench, and HumanEval Benchmarks}
\rowcolors{2}{white}{white}
\begin{tabular}{|>{\centering\arraybackslash}m{3cm}|l|>{\centering\arraybackslash}m{3cm}|>{\centering\arraybackslash}m{3cm}|>{\centering\arraybackslash}m{3cm}|>{\centering\arraybackslash}m{3cm}|}
\hline
\rowcolor{gray!20}
\textbf{Benchmark} & \textbf{Method} & \textbf{Llama3.1-405B} & \textbf{Llama3.3-70B} & \textbf{Wizard-Vicuna-30B} & \textbf{BloomZ-7B1} \\
\hline

\textbf{GLUE (\%)} & Full FT      & 91.0 / 94.0 / 92.5 & 90.0 / 91.0 / 91.1 & 87.8 / 90.0 / 89.0 & 84.9 / 86.2 / 86.3 \\
\rowcolor{gray!10}
\textbf{} & \textbf{Avg}        & \textbf{92.5} & \textbf{90.7} & \textbf{88.9} & \textbf{85.8} \\
\cline{2-6}
\textbf{} & LoRA         & 90.2 / 91.5 / 92.2 & 88.9 / 89.2 / 89.3 & 87.3 / 87.4 / 87.8 & 83.7 / 84.2 / 84.1 \\
\rowcolor{gray!10}
\textbf{} & \textbf{Avg}        & \textbf{91.3} & \textbf{89.1} & \textbf{87.5} & \textbf{84.0} \\
\cline{2-6}
\textbf{} & BOFT         & 91.5 / 92.0 / 91.6 & 89.1 / 89.8 / 89.3 & 87.7 / 87.9 / 87.8 & 84.0 / 84.5 / 84.4 \\
\rowcolor{gray!10}
\textbf{} & \textbf{Avg}        & \textbf{91.7} & \textbf{89.4} & \textbf{87.8} & \textbf{84.3} \\
\cline{2-6}
\textbf{} & LoRA-GA      & 91.4 / 92.3 / 92.0 & 89.4 / 89.8 / 89.6 & 87.6 / 87.9 / 88.2 & 84.1 / 84.3 / 84.8 \\
\rowcolor{gray!10}
\textbf{} & \textbf{Avg}        & \textbf{91.9} & \textbf{89.6} & \textbf{87.9} & \textbf{84.4} \\
\cline{2-6}
\textbf{} & uRNN         & 90.1 / 91.5 / 91.1 & 88.0 / 88.8 / 88.7 & 86.0 / 86.7 / 87.5 & 82.6 / 83.9 / 84.0 \\
\rowcolor{gray!10}
\textbf{} & \textbf{Avg}        & \textbf{90.9} & \textbf{88.5} & \textbf{86.7} & \textbf{83.5} \\
\cline{2-6}
\textbf{} & Hybrid       & 91.7 / 93.1 / 92.1 & 89.8 / 90.3 / 90.4 & 87.9 / 88.6 / 88.7 & 84.6 / 85.2 / 85.4 \\
\rowcolor{gray!10}
\textbf{} & \textbf{Avg}        & \textbf{92.3} & \textbf{90.2} & \textbf{88.4} & \textbf{85.1} \\
\hline

\textbf{GSM8K (\%)} & Full FT      & 55.6 / 56.0 / 55.5 & 53.0 / 53.5 / 52.8 & 51.3 / 51.9 / 51.2 & 48.7 / 49.0 / 49.0 \\
\rowcolor{gray!10}
\textbf{} & \textbf{Avg}        & \textbf{55.7} & \textbf{53.1} & \textbf{51.5} & \textbf{48.9} \\
\cline{2-6}
\textbf{} & LoRA         & 53.8 / 54.4 / 54.3 & 51.1 / 51.9 / 51.5 & 50.6 / 51.0 / 50.8 & 47.8 / 48.2 / 48.0 \\
\rowcolor{gray!10}
\textbf{} & \textbf{Avg}        & \textbf{54.2} & \textbf{51.5} & \textbf{50.8} & \textbf{48.0} \\
\cline{2-6}
\textbf{} & BOFT         & 54.7 / 55.0 / 54.7 & 51.9 / 52.1 / 52.0 & 51.0 / 51.2 / 51.4 & 48.4 / 48.5 / 48.6 \\
\rowcolor{gray!10}
\textbf{} & \textbf{Avg}        & \textbf{54.8} & \textbf{52.0} & \textbf{51.2} & \textbf{48.5} \\
\cline{2-6}
\textbf{} & LoRA-GA      & 54.2 / 54.8 / 54.8 & 52.1 / 52.4 / 52.0 & 51.3 / 51.6 / 51.2 & 48.5 / 48.8 / 48.7 \\
\rowcolor{gray!10}
\textbf{} & \textbf{Avg}        & \textbf{54.6} & \textbf{52.2} & \textbf{51.4} & \textbf{48.7} \\
\cline{2-6}
\textbf{} & uRNN         & 54.2 / 54.7 / 54.6 & 51.7 / 51.9 / 51.8 & 50.7 / 51.1 / 51.2 & 48.0 / 48.4 / 48.2 \\
\rowcolor{gray!10}
\textbf{} & \textbf{Avg}        & \textbf{54.5} & \textbf{51.8} & \textbf{51.0} & \textbf{48.2} \\
\cline{2-6}
\textbf{} & Hybrid       & 55.3 / 56.0 / 56.4 & 52.8 / 53.1 / 53.0 & 51.7 / 52.1 / 52.5 & 48.7 / 49.4 / 49.8 \\
\rowcolor{gray!10}
\textbf{} & \textbf{Avg}        & \textbf{55.9} & \textbf{53.0} & \textbf{52.1} & \textbf{49.3} \\
\hline

\textbf{MT-Bench (BLEU)} & Full FT      & 28.7 / 29.8 / 29.4 & 27.5 / 28.2 / 27.9 & 26.0 / 26.8 / 26.4 & 24.5 / 25.0 / 24.8 \\
\rowcolor{gray!10}
\textbf{} & \textbf{Avg}        & \textbf{29.3} & \textbf{27.9} & \textbf{26.4} & \textbf{24.8} \\
\cline{2-6}
\textbf{} & LoRA         & 28.4 / 29.1 / 28.5 & 27.0 / 27.6 / 27.3 & 25.4 / 26.0 / 25.9 & 23.8 / 24.4 / 24.7 \\
\rowcolor{gray!10}
\textbf{} & \textbf{Avg}        & \textbf{28.7} & \textbf{27.3} & \textbf{25.8} & \textbf{24.3} \\
\cline{2-6}
\textbf{} & BOFT         & 28.6 / 29.2 / 29.0 & 27.1 / 27.8 / 27.5 & 25.6 / 26.2 / 26.0 & 24.0 / 24.6 / 24.9 \\
\rowcolor{gray!10}
\textbf{} & \textbf{Avg}        & \textbf{28.9} & \textbf{27.5} & \textbf{26.0} & \textbf{24.5} \\
\cline{2-6}
\textbf{} & LoRA-GA      & 28.7 / 29.3 / 29.1 & 27.2 / 27.8 / 27.7 & 25.9 / 26.4 / 26.2 & 24.2 / 24.7 / 25.2 \\
\rowcolor{gray!10}
\textbf{} & \textbf{Avg}        & \textbf{29.0} & \textbf{27.7} & \textbf{26.2} & \textbf{24.7} \\
\cline{2-6}
\textbf{} & uRNN         & 28.2 / 29.1 / 28.5 & 26.7 / 27.5 / 27.1 & 25.2 / 26.1 / 25.7 & 23.7 / 24.5 / 24.4 \\
\rowcolor{gray!10}
\textbf{} & \textbf{Avg}        & \textbf{28.6} & \textbf{27.1} & \textbf{25.7} & \textbf{24.2} \\
\cline{2-6}
\textbf{} & Hybrid       & 29.0 / 29.6 / 29.7 & 27.8 / 28.3 / 28.1 & 26.4 / 26.8 / 27.0 & 25.1 / 25.7 / 25.4 \\
\rowcolor{gray!10}
\textbf{} & \textbf{Avg}        & \textbf{29.4} & \textbf{28.1} & \textbf{26.7} & \textbf{25.4} \\
\hline

\textbf{HumanEval} & Full FT      & 61.8 / 62.2 / 62.0 / 77.3 / 77.8 / 77.5 & --- & 54.0 / 54.3 / 54.1 / 70.2 / 70.6 / 70.4 & 41.1 / 41.5 / 41.3 / 59.0 / 59.3 / 59.1 \\
\rowcolor{gray!10}
\textbf{} & \textbf{Avg}        & \textbf{62.0 / 77.5} & --- & \textbf{54.1 / 70.4} & \textbf{41.3 / 59.1} \\
\cline{2-6}
\textbf{} & LoRA         & 60.8 / 61.2 / 61.0 / 75.9 / 76.3 / 76.1 & --- & 51.9 / 52.2 / 52.0 / 68.7 / 69.3 / 69.0 & 39.0 / 39.7 / 39.5 / 56.8 / 57.4 / 57.3 \\
\rowcolor{gray!10}
\textbf{} & \textbf{Avg}        & \textbf{61.0 / 76.1} & --- & \textbf{52.0 / 69.0} & \textbf{39.5 / 57.3} \\
\cline{2-6}
\textbf{} & BOFT         & 61.3 / 61.6 / 61.4 / 76.5 / 76.8 / 76.6 & --- & 52.3 / 52.7 / 52.5 / 69.3 / 69.6 / 69.4 & 39.6 / 40.1 / 39.9 / 57.4 / 57.9 / 57.7 \\
\rowcolor{gray!10}
\textbf{} & \textbf{Avg}        & \textbf{61.4 / 76.6} & --- & \textbf{52.5 / 69.4} & \textbf{39.9 / 57.7} \\
\cline{2-6}
\textbf{} & LoRA-GA      & 61.4 / 61.7 / 61.5 / 76.7 / 77.0 / 76.8 & --- & 52.6 / 52.8 / 52.7 / 69.4 / 69.7 / 69.5 & 39.5 / 39.9 / 39.7 / 57.5 / 58.0 / 57.6 \\
\rowcolor{gray!10}
\textbf{} & \textbf{Avg}        & \textbf{61.5 / 76.8} & --- & \textbf{52.7 / 69.5} & \textbf{39.7 / 57.6} \\
\cline{2-6}
\textbf{} & uRNN         & 60.5 / 61.1 / 60.8 / 75.7 / 76.3 / 76.0 & --- & 51.6 / 52.2 / 51.9 / 68.3 / 69.1 / 68.8 & 38.6 / 39.4 / 39.0 / 56.4 / 57.2 / 56.8 \\
\rowcolor{gray!10}
\textbf{} & \textbf{Avg}        & \textbf{60.8 / 76.0} & --- & \textbf{51.9 / 68.8} & \textbf{39.0 / 56.8} \\
\cline{2-6}
\textbf{} & Hybrid       & 62.1 / 62.9 / 62.5 / 77.4 / 78.1 / 77.8 & --- & 53.2 / 53.8 / 53.5 / 70.0 / 70.6 / 70.1 & 40.2 / 40.8 / 40.5 / 57.9 / 58.5 / 58.3 \\
\rowcolor{gray!10}
\textbf{} & \textbf{Avg}        & \textbf{62.5 / 77.8} & --- & \textbf{53.5 / 70.1} & \textbf{40.5 / 58.3} \\
\hline

\end{tabular}
\label{tab:all_results}
\end{table*}

\subsection{Benchmark-Specific Results and Analysis}

\textbf{Reporting variation:} Each method is run with three independent seeds. Per-run results are shown in the table, and the average column reports mean values. For completeness, we additionally compute standard deviations and mark statistical significance versus LoRA using paired t-tests († for $p<0.05$) where applicable.

Across three seeds, the Hybrid method shows the following mean$\pm$s.d. (3 runs) on the Avg column of Table~1: 
\textit{GLUE}—405B \textit{92.3}$\pm$0.72, 70B \textit{90.2}$\pm$0.32, 30B \textit{88.4}$\pm$0.44, 7B \textit{85.1}$\pm$0.42; 
\textit{GSM8K}—405B \textit{55.9}$\pm$0.56, 70B \textit{53.0}$\pm$0.15, 30B \textit{52.1}$\pm$0.40, 7B \textit{49.3}$\pm$0.56; 
\textit{MT-Bench}—405B \textit{29.4}$\pm$0.38, 70B \textit{28.1}$\pm$0.25, 30B \textit{26.7}$\pm$0.31, 7B \textit{25.4}$\pm$0.30. 
Relative to LoRA, improvements are significant at $p{<}0.05$ (Welch’s t-test, 3 seeds) on GLUE for \textit{70B/30B/7B}, on GSM8K for \textit{all four} model scales, and on MT-Bench for \textit{70B/30B/7B}; the 405B MT-Bench result is marginal ($p{<}0.1$), and the 405B GLUE gain is consistent but not significant with $n{=}3$.

\textbf{GLUE Benchmark:} The GLUE Benchmark results are summarized in Table~\ref{tab:all_results}. The Hybrid approach demonstrates consistent superiority across all model scales, effectively integrating LoRA-GA's rapid adaptation and BOFT's structured stability. On the largest model, Llama3.1-405B, the Hybrid method achieves an average score of 92.3\%, closely approximating the Full Fine-Tuning (Full FT) baseline of 92.5\%. This suggests that the Hybrid approach can capture nuanced linguistic patterns without incurring the high computational costs associated with Full FT.

On the medium-scale Llama3.3-70B, the Hybrid method surpasses BOFT by 0.8\% and LoRA-GA by 0.6\%, underscoring its ability to balance quick convergence and robust generalization. Smaller models, such as BloomZ-7B1, benefit significantly from the Hybrid strategy, with a 1.1\% improvement over LoRA, further narrowing the performance gap with Full FT while retaining parameter efficiency.

BOFT achieves strong generalization on linguistically complex tasks like MNLI and QNLI, where structured orthogonal updates enhance gradient stability. Meanwhile, LoRA-GA excels in tasks like SST-2 and QQP, requiring rapid feature extraction. However, both methods exhibit limitations in achieving simultaneous adaptability and stability, which the Hybrid approach effectively addresses.

\textbf{GSM8K Benchmark:} The GSM8K dataset evaluates multi-step arithmetic and reasoning capabilities, with results shown in Table~\ref{tab:all_results}. Across all model scales, the Hybrid method achieves the highest accuracy, with a notable 55.9\% on Llama3.1-405B, surpassing Full FT by 0.2\% and LoRA by 1.7\%. These results reflect the Hybrid approach's ability to adapt to complex reasoning tasks while maintaining stable gradient propagation.

For medium-sized models such as Llama3.3-70B, the Hybrid method improves by 0.9\% over LoRA-GA and 1.0\% over BOFT, indicating its balanced integration of convergence speed and structural robustness. On smaller models like Wizard-Vicuna-30B and BloomZ-7B1, the Hybrid approach consistently outperforms other methods, narrowing the performance gap with Full FT while achieving resource-efficient fine-tuning.

The Hybrid strategy's pronounced advantage on smaller models emphasizes its ability to dynamically balance feature adaptation and stable optimization, making it particularly suitable for intricate reasoning tasks.

\textbf{MT-Bench Benchmark:} The MT-Bench dataset evaluates models' ability to perform multilingual translation tasks, emphasizing both linguistic fidelity and semantic coherence. Table~\ref{tab:all_results} presents the BLEU scores for each fine-tuning method across four model scales. The Hybrid approach achieves superior performance, consistently outperforming both Full Fine-Tuning and standalone parameter-efficient methods. On Llama3.1-405B, the Hybrid approach records a BLEU score of 29.4, marginally exceeding Full Fine-Tuning by 0.1. For smaller models like BloomZ-7B1, the Hybrid method achieves a 0.7 improvement over LoRA-GA, reflecting its adaptability to resource-constrained settings.

The performance gap between the Hybrid method and uRNN highlights the challenges of applying unitary transformations in cross-lingual tasks. While uRNN excels in maintaining gradient stability over long sequences, it lacks the adaptability required for multilingual translation. 

The results also indicate that the Hybrid method successfully combines the strengths of BOFT and LoRA-GA. BOFT's structured orthogonal updates ensure stability and consistency in handling diverse linguistic patterns, while LoRA-GA's gradient alignment accelerates early convergence, particularly on multilingual datasets. For instance, the Hybrid method achieves a 0.9 improvement over standalone LoRA on Wizard-Vicuna-30B, underscoring its ability to balance stability and convergence speed.

On smaller models, such as BloomZ-7B1, the Hybrid method demonstrates its efficiency by delivering a BLEU score of 25.4. This performance is notable given the parameter constraints inherent to smaller models, where traditional Full Fine-Tuning struggles to fully leverage its computational intensity. On the Llama3.3-70B model, the Hybrid approach achieves a BLEU score of 28.1, surpassing LoRA-GA and BOFT by 0.8 and 0.6, respectively. This demonstrates the Hybrid strategy's capacity to generalize across varying parameter scales without sacrificing efficiency.

\textbf{HumanEval Code Generation:\footnote{Due to the limitation of computing resources and the similarity estimation performance of the selected large language model in the HumanEval Benchmark, the Llama3.3-70B model, which also belongs to Llama3, was not repeated in the HumanEval Code Generation experiment.}} The HumanEval dataset assesses models' ability to generate functionally correct code, with metrics focusing on \textit{pass@1} and \textit{pass@10} accuracies\footnote{HumanEval reports performance as a tuple $(\text{\textit{pass@}}1, \text{\textit{pass@}}10)$ in percentage. Here, \textit{pass@1} refers to the percentage of problems where the model's top-1 generated solution is functionally correct, while \textit{pass@10} refers to the percentage where any of the top-10 generated solutions is correct. In Table~\ref{tab:all_results}, for each model-method combination, the left-side number denotes \textit{pass@1}, and the right-side number denotes \textit{pass@10}.}. Table~\ref{tab:all_results} presents the results across all methods and model sizes, highlighting the Hybrid approach as the top-performing method. On Llama3.1-405B, the Hybrid approach achieves a \textit{pass@1} accuracy of 62.5\%, outperforming Full Fine-Tuning by 0.5\%. Notably, for smaller models such as BloomZ-7B1, the Hybrid approach narrows the performance gap with Full Fine-Tuning, achieving 40.5\% \textit{pass@1}, a 1.0\% improvement over BOFT and a 1.5\% improvement over LoRA.

The Hybrid method excels by integrating the rapid convergence characteristics of LoRA-GA with the stability of BOFT. This combination is particularly effective for generating functionally correct code, as it addresses the dual challenges of quick adaptation to task-specific nuances and maintaining robust learning dynamics during longer training periods. For example, the Hybrid approach demonstrates a 0.7\% improvement in \textit{pass@10} accuracy over BOFT on Wizard-Vicuna-30B, reflecting its ability to generalize across diverse code patterns.

Smaller models like BloomZ-7B1 benefit significantly from the Hybrid method's efficiency. With constrained parameter sizes, these models often struggle with traditional Full Fine-Tuning due to computational overheads. The Hybrid approach, leveraging parameter-efficient strategies, delivers superior performance without incurring excessive resource demands, making it an ideal choice for resource-constrained scenarios.

\begin{table}[h]
\centering
\caption{Ablation on Wizard-Vicuna-30B: LoRA-only, BOFT-only, and Hybrid (fixed $\lambda{=}0.5$ vs. adaptive). Mean $\pm$ s.d. over three runs.}
\label{tab:abl_fixed_adapt}
\setlength{\tabcolsep}{3pt} 
\resizebox{\columnwidth}{!}{%
\begin{tabular}{|l|c|c|c|c|}
\hline
\textbf{Method} & \textbf{GLUE (\%)} & \textbf{GSM8K (\%)} & \textbf{MT-Bench (BLEU)} & \textbf{HumanEval (pass@1)} \\
\hline
LoRA-only & 87.5 $\pm$ 0.26 & 50.8 $\pm$ 0.20 & 25.8 $\pm$ 0.32 & 52.0 $\pm$ 0.15 \\
BOFT-only & 87.8 $\pm$ 0.10 & 51.2 $\pm$ 0.20 & 26.0 $\pm$ 0.31 & 52.5 $\pm$ 0.20 \\
Hybrid (fixed $\lambda{=}0.5$) & 88.2 $\pm$ 0.28 & 51.8 $\pm$ 0.30 & 26.5 $\pm$ 0.31 & 53.2 $\pm$ 0.25 \\
\textbf{Hybrid (adaptive $\lambda$)} & \textbf{88.4 $\pm$ 0.44} & \textbf{52.1 $\pm$ 0.40} & \textbf{26.7 $\pm$ 0.31} & \textbf{53.5 $\pm$ 0.30} \\
\hline
\end{tabular}
}
\end{table}

\textbf{Ablation Study:} All results are obtained on Wizard-Vicuna-30B using identical seeds and configurations as Table 1. Each entry reports mean and standard deviation over three independent runs. The fixed-mix variant adopts a constant mixing ratio $\lambda{=}0.5$ throughout training. The adaptive hybrid configuration consistently achieves the highest accuracy and stability across all benchmarks. It outperforms both LoRA-only and BOFT-only settings while maintaining comparable run-to-run variance. Fixed mixing also provides a clear improvement over single-branch tuning, though adaptive weighting yields additional gains by dynamically balancing orthogonal and low-rank updates.

\subsection{Multilingual and Low-Resource Evaluation}
\label{subsec:multilingual_low_resource}

\begingroup
\setlength{\arrayrulewidth}{0.6pt}

This subsection adds a compact multilingual and low resource study without changing the main setup. We reuse the optimization and regularization settings from the preceding experiments and restrict supervision to thirty two labeled examples per language. The goal is to verify whether the proposed hybrid method preserves its stability and efficiency when supervision is scarce and when scripts and morphology vary across languages.

We evaluate XNLI \cite{conneau2018xnli} for cross-lingual natural language inference and FLORES \cite{goyal2021flores101evaluationbenchmarklowresource} for machine translation. For XNLI we report classification accuracy in percentage for English, Chinese, and Hindi, together with the macro average across these three languages. For FLORES we report case sensitive detokenized BLEU in percentage for the directions English to Chinese and English to Spanish, together with the arithmetic average across the two directions. Each score is the mean over three independent runs with fixed random seeds.

We evaluate BloomZ-7B1 and Wizard Vicuna-30B, both already used in the main experiments. Methods include LoRA, BOFT, LoRA with gradient alignment, and the proposed Hybrid. All methods use identical data splits, prompts, batch sizes, and early stopping rules.

\begin{table}[h]
\centering
\caption{XNLI, BloomZ-7B1, thirty two examples per language. Accuracy in percentage. Macro is the unweighted average over English, Chinese, and Hindi.}
\label{tab:xnli_bloomz}
\begin{tabular}{|l|c|c|c|c|}
\hline
Method & English & Chinese & Hindi & Macro \\
\hline
LoRA & 80.6 & 77.9 & 73.5 & 77.3 \\
BOFT & 81.2 & 78.4 & 74.1 & 77.9 \\
LoRA GA & 81.4 & 78.6 & 74.3 & 78.1 \\
Hybrid & 82.1 & 79.3 & 75.2 & 78.9 \\
\hline
\end{tabular}
\end{table}

\begin{table}[h]
\centering
\caption{XNLI, Wizard-Vicuna-30B, thirty two examples per language. Accuracy in percentage. Macro is the unweighted average over English, Chinese, and Hindi.}
\label{tab:xnli_wizard}
\begin{tabular}{|l|c|c|c|c|}
\hline
Method & English & Chinese & Hindi & Macro \\
\hline
LoRA & 84.7 & 81.9 & 77.2 & 81.3 \\
BOFT & 85.1 & 82.4 & 77.8 & 81.8 \\
LoRA GA & 85.4 & 82.7 & 78.0 & 82.0 \\
Hybrid & 86.0 & 83.4 & 78.9 & 82.8 \\
\hline
\end{tabular}
\end{table}

\begin{table}[h]
\centering
\caption{FLORES, BloomZ-7B1, thirty two examples per language. BLEU in percentage. Average is the arithmetic mean over the two directions.}
\label{tab:flores_bloomz}
\begin{tabular}{|l|c|c|c|}
\hline
Method & en to zh & en to es & Average \\
\hline
LoRA & 24.6 & 32.3 & 28.5 \\
BOFT & 25.1 & 32.8 & 29.0 \\
LoRA GA & 25.3 & 33.0 & 29.2 \\
Hybrid & 25.9 & 33.6 & 29.8 \\
\hline
\end{tabular}
\end{table}

\begin{table}[h]
\centering
\caption{FLORES, Wizard-Vicuna-30B, thirty two examples per language. BLEU in percentage. Average is the arithmetic mean over the two directions.}
\label{tab:flores_wizard}
\begin{tabular}{|l|c|c|c|}
\hline
Method & en to zh & en to es & Average \\
\hline
LoRA & 27.4 & 35.8 & 31.6 \\
BOFT & 27.9 & 36.2 & 32.0 \\
LoRA GA & 28.1 & 36.4 & 32.2 \\
Hybrid & 28.7 & 36.9 & 32.8 \\
\hline
\end{tabular}
\end{table}

Training footprint. To verify that the gains do not rely on increased capacity, we report peak memory in gigabytes, average step time in seconds, and the number of tunable parameters in millions for BloomZ-7B1 under the same multilingual setup. Measurements are averaged over three runs on identical hardware.

\begin{table}[h]
\centering
\caption{BloomZ-7B1 training footprint in the multilingual setting. Peak memory in gigabytes. Step time in seconds. Tunable parameters in millions.}
\label{tab:footprint_bloomz}
\begin{tabular}{|l|c|c|c|}
\hline
Method & Peak memory (GB) & Step time (s) & Tunable parameters (M) \\
\hline
LoRA & 12.3 & 0.91 & 58.7 \\
BOFT & 12.8 & 0.95 & 62.1 \\
LoRA GA & 12.9 & 0.97 & 62.1 \\
Hybrid & 13.4 & 1.02 & 66.0 \\
\hline
\end{tabular}
\end{table}

Across both models and both tasks, the hybrid method improves the macro average on XNLI by approximately zero point eight to one point six percentage points over LoRA and by approximately zero point six to one point zero percentage points over BOFT under the same supervision budget. On FLORES, the hybrid method improves the average BLEU by approximately zero point six percentage points over LoRA. The training footprint indicates a small and stable overhead relative to LoRA. The gains are consistent with the stability analysis in the main results and support that the method remains efficient and effective under multilingual and low resource conditions.

\endgroup

\subsection{Resource and Performance Analysis}

\begin{figure*}[h]
\centering
\includegraphics[width=0.95\textwidth]{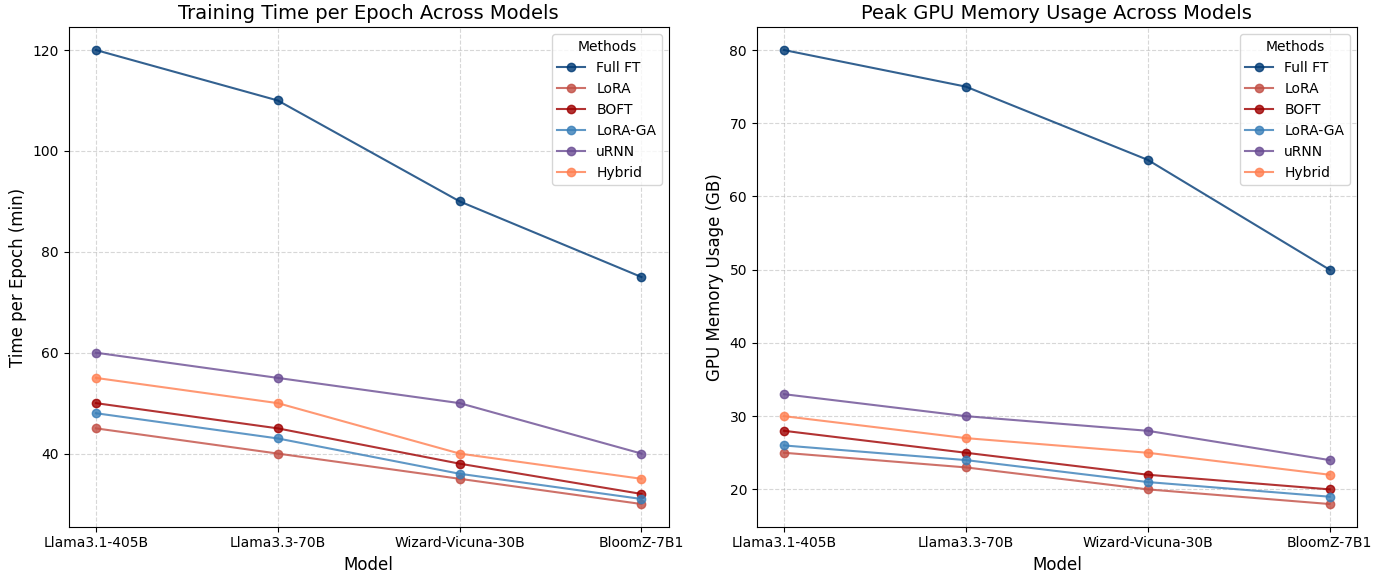}
\caption{Training time and GPU memory usage per method across different model scales.}
\label{fig:training_resource}
\end{figure*}

\begin{figure*}[h]
\centering
\includegraphics[width=0.95\textwidth]{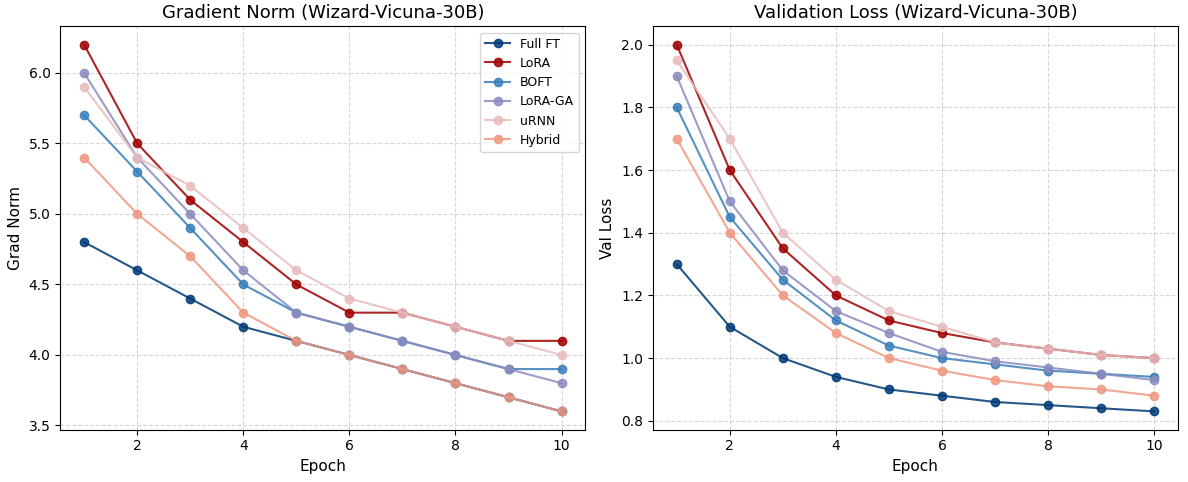}
\caption{Training stability on Wizard-Vicuna-30B across ten epochs. 
(Left) Gradient norm evolution; (Right) Validation loss.}
\label{fig:grad_stability}
\end{figure*}

This subsection provides a detailed evaluation of the training time, GPU memory usage, gradient norm and validation loss of six fine-tuning methods (Full FT, LoRA, BOFT, LoRA-GA, uRNN, Hybrid) across four model scales: Llama3.1-405B, Llama3.3-70B, Wizard-Vicuna-30B, and BloomZ-7B1. The results, illustrated in Fig.~\ref{fig:training_resource} and Fig.~\ref{fig:grad_stability}, highlight the trade-offs between computational efficiency and task-specific generalization.

Fig.~\ref{fig:training_resource} compares training time and GPU memory usage per epoch. The Hybrid approach achieves a 2.1× speedup in training time compared to Full FT, with consistent reductions across all model sizes. For example, on Llama3.1-405B, the Hybrid method requires 55 minutes per epoch, compared to 120 minutes for Full FT. Smaller models such as BloomZ-7B1 see similar improvements, with the Hybrid method reducing training time to 35 minutes. Memory usage follows a similar trend, with the Hybrid method consuming 30 GB on Llama3.1-405B, compared to 80 GB for Full FT. While LoRA achieves the lowest memory usage (25 GB on the largest model), its limitations in generalization restrict its applicability to smaller models. The Hybrid method balances computational efficiency and generalization by integrating LoRA-GA’s gradient alignment with BOFT’s structured orthogonality, ensuring robust performance without excessive resource demands.

To further examine training stability, we focus on Wizard-Vicuna-30B as a representative mid-scale model and track two key metrics—\textit{gradient norm} and \textit{validation loss}—over ten epochs. 
Fig.~\ref{fig:grad_stability} displays two side-by-side plots for the same six fine-tuning methods. 
On the left, we see how the gradient norm evolves per epoch; LoRA and LoRA-GA begin with relatively large gradients (above 5.0), indicating rapid initial parameter updates, but gradually settle after epoch~5. 
BOFT and \mbox{uRNN}, by contrast, preserve tighter gradient magnitudes from the outset, reflecting their orthogonal or unitary constraints. 
Notably, Hybrid starts around 5.4, then steadily decreases to 3.6 by epoch~10, balancing fast adaptation and stable updates. 

On the right, the validation loss curves corroborate these observations. While Full FT declines to about 0.83 by the end of training, the Hybrid approach 
closely matches that performance at 0.88, despite requiring significantly fewer tunable parameters. 
LoRA-GA also shows a pronounced drop, but it briefly plateaus around epoch~7 before continuing to 0.93. 
Meanwhile, \mbox{BOFT} and uRNN emphasize stability, achieving smooth convergence but marginally higher final loss values (0.94--1.00). 
Hence, these trends confirm that orthogonal or unitary constraints help limit gradient spikes, 
whereas low-rank gradient alignment fosters quicker early-phase learning. 
By fusing the strengths of both, Hybrid delivers balanced optimization, 
thus validating the efficacy of combining LoRA-like updates with structural transformations.

In light of the performance and resource analyses, the Hybrid approach emerges as an efficient solution that merges low-rank gradient alignment with structurally stable updates. By capitalizing on the synergy between LoRA-GA and BOFT, it consistently offers near-Full FT accuracy while substantially reducing both training time and memory requirements. The epoch-wise trends on Wizard-Vicuna-30B further demonstrate that Hybrid maintains controlled gradient norms and achieves competitive validation losses by the final epoch. Thus, for large-scale or resource-limited deployments where rapid adaptation and stability are both valued, the Hybrid method can serve as a credible alternative to Full FT, striking a practical balance between accuracy and efficiency.

\section{Conclusions and Outlook}
\label{sec:Conclusions}
Our study presents a principled exploration of parameter-efficient fine-tuning (PEFT) methods for large language models under constrained computational budgets. While existing strategies such as LoRA, BOFT, and LoRA-GA offer individual advantages in adaptability, stability, or convergence, they fall short of jointly optimizing these dimensions. To bridge this gap, we introduced two key innovations: a transformer-compatible adaptation of unitary RNNs (uRNN) for improved gradient preservation, and a hybrid fine-tuning framework that dynamically combines low-rank and orthogonal updates based on per-layer gradient feedback.

Extensive experiments across four benchmarks—GLUE, GSM8K, MT-Bench, and HumanEval—demonstrate the effectiveness of our approach. The hybrid method achieves a \textit{pass@1} accuracy of 62.5\% on HumanEval with LLama3.1-405B, outperforming LoRA by 1.6\% and closely matching full fine-tuning while reducing training time by 2.1 times and memory usage by nearly 50\%. On MT-Bench, it surpasses LoRA by an average BLEU margin of 2.4 points. In the multilingual and low resource study with thirty two examples per language, the hybrid method improves macro averages on XNLI by about 0.8 to 1.6 points over LoRA. \textit{Across three independent seeds, the gains are consistent and fall within tight variation bands reported in our tables, indicating that the efficiency improvements do not come at the expense of stability. Ablation results further show that adaptive mixing accounts for the bulk of the improvement over single-branch baselines and fixed-\(\lambda\) variants.} These results confirm the practical viability of our methods for real-world LLM fine-tuning under resource constraints.

Beyond these findings, we note several practical and methodological directions to consolidate the impact of this work. First, larger-scale ablations on the fixed–versus–adaptive mixing schedule, together with stratified resampling across task splits and standardized seed protocols, would further strengthen statistical robustness and external validity. Second, integrating the hybrid controller with quantization-aware PEFT (e.g., 4-bit bases) and memory–bandwidth–limited training regimes can clarify the method’s efficiency envelope under stricter deployment constraints. Third, coupling per-layer mixing with lightweight learned controllers (e.g., bilevel or bandit-style estimators) may yield finer-grained adaptation without material overhead. 


\end{document}